\documentclass[runningheads]{llncs}
\usepackage{graphicx}
\usepackage{comment}
\usepackage{amsmath,amssymb} 
\usepackage{color}
\usepackage{mathtools}
\usepackage{caption}
\usepackage{subcaption}
\usepackage{booktabs}
\usepackage{multirow}
\usepackage{epsfig}
\usepackage{multirow}
\usepackage{csquotes}
\usepackage{tikz}

\begin{document}
\pagestyle{headings}
\mainmatter
\def\ECCVSubNumber{2202}  

\title{CAFE-GAN: Arbitrary Face Attribute Editing\\ with Complementary Attention Feature} 


\titlerunning{CAFE-GAN}
%
\author{Jeong-gi Kwak\inst{1} \and
David K. Han\inst{2} \and
Hanseok Ko\inst{1}}
\authorrunning{J. Kwak et al.}
%
\institute{Korea University, Seoul, Korea \and
Army Research Laboratory, Adelphi, MD, USA}
\maketitle

\begin{abstract}
The goal of face attribute editing is altering a facial image according to given target attributes such as hair color, mustache, gender, etc. 
It belongs to the image-to-image domain transfer problem with a set of attributes considered as a distinctive domain.
There have been some works in multi-domain transfer problem focusing on facial attribute editing employing Generative Adversarial Network (GAN). These methods have reported some successes but they also result in unintended changes in facial regions - meaning the generator alters regions unrelated to the specified attributes.
To address this unintended altering problem, we propose a novel GAN model which is designed to edit only the parts of a face pertinent to the target attributes by the concept of Complementary Attention Feature (CAFE).
CAFE identifies the facial regions to be transformed by considering both target attributes as well as \enquote{complementary attributes}, which we define as those attributes absent in the input facial image. In addition, we introduce a complementary feature matching to help in training the generator for utilizing the spatial information of attributes.
Effectiveness of the proposed method is demonstrated by analysis and comparison study with state-of-the-art methods.

\keywords{Face Attribute Editing, GAN, Complementary Attention Feature, Complementary Feature Matching}
\end{abstract}

\section{Introduction}

Since the advent of GAN by Goodfellow \textit{et al.}~\cite{goodfellow2014generative}, its application literally exploded into a variety of areas, and many variants of GAN emerged. 
Conditional GAN (CGAN)~\cite{mirza2014conditional}, one of the GAN variants, adds an input as a condition on how the synthetic output should be generated.
An area in CGAN receiving a particular attention in the media is \enquote{Deep Fake} in which an input image is transformed into a new image of different nature while key elements of the original image are retained and transposed~\cite{Isola_2017_CVPR,mirza2014conditional}. 
GAN based style transfer is often the method of choice for achieving the domain transfer of the input to another domain in the output, such as generating a hypothetical painting of a well known artist from a photograph.  
CycleGAN~\cite{Zhu_2017} has become a popular method in image domain transfer, because it uses cycle-consistency loss from a single image input in training, and thus its training is unsupervised.

Nevertheless, single domain translation models~\cite{Shen_2017,Zhang_2018_ECCV} including CycleGAN are incapable of learning domain transfer to multiple domains.  
Thus, in these approaches, multiple models are required to match the number of target domains. 
One of the multi-domain transfer problems is manipulation of one's facial characteristics.
The goal of facial attribute editing is to convert specific attributes of a face, such as wearing eyeglasses, to a face without eyeglasses. 
Other attributes may include local properties e.g., beard as well as global properties e.g., face aging. 
Obviously this requires multiples of domain transfer models if a single domain transfer concept is to be used.  
The number of single domain transfer models required in such a case is a function of the attribute combination since these facial attributes are mostly independent. Even for relatively small number of attributes, single domain transfer approaches would require a significantly high number of separate models. 
A model such as CycleGAN, therefore, would become impractical. 
\begin{figure}[!t]
\centering \includegraphics[width=\linewidth]{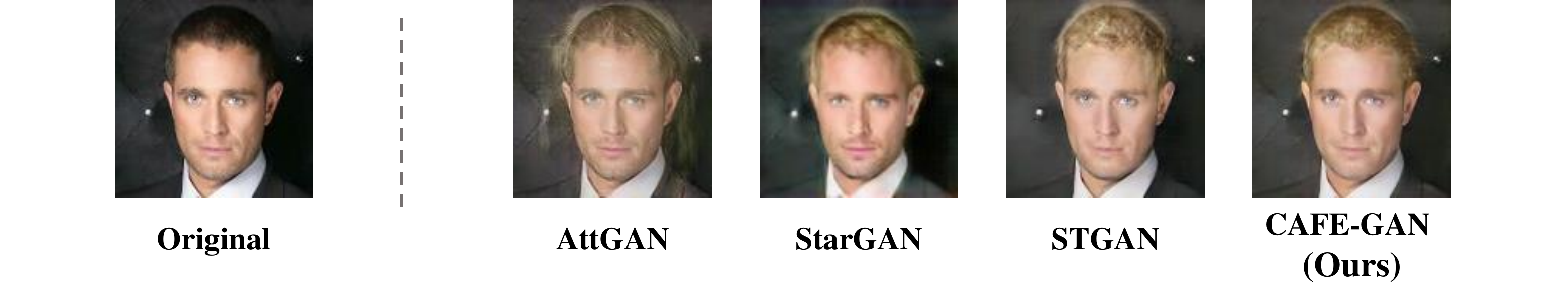}
\caption{Face editing results of AttGAN~\cite{He2017AttGANFA}, StarGAN~\cite{Choi_2018}, STGAN~\cite{Liu_2019_CVPR} and our model given a target attribute \textit{Blond hair}. While AttGAN, StarGAN and STGAN deliver blond hair, they also create unwanted changes (e.g. halo, different hair style, etc.) in resultant images.}
\label{fig:f1}
\end{figure}  
To address the multi-domain transfer problem, StarGAN~\cite{Choi_2018} and AttGAN~\cite{He2017AttGANFA}, have been proposed by introducing a target vector of multiple attributes as an additional input.
These target attribute vector based GAN models have shown some impressive images, but, they often result in unintended changes - meaning the generator alters regions unrelated to specified attributes as shown in Fig.~\ref{fig:f1}. It stems from these models driven to achieve higher objective scores in classifying attributes at the expense of affecting areas unrelated to the target attribute. Some methods~\cite{Shen_2017,Zhang_2018_ECCV} have used a strategy that adds only the values of attribute-specific region to an input image, but these methods have achieved limited success.  Hence, changing only pertinent regions remains as an important but challenging problem. Their limitation stems from considering only structural improvement in the generator. However, more effective approach may be possible by exploring decision making process of the discriminator.

Visual explanation~\cite{fukui2019attention,selvaraju2017grad,springenberg2014striving,zhou2016learning}, which is known effective in interpreting Convolutional Neural Network (CNN) by highlighting response areas critical in recognition, is considered here to address the problem. Our model is mainly motivated by Attention Branch Network (ABN)~\cite{fukui2019attention} which extends response-based visual explanation to an attention mechanism. In ABN, the attention branch takes mid-level features then extracts attention feature maps. Attention feature maps are then downsampled through a global average pooling layer (GAP)~\cite{DBLP:journals/corr/LinCY13} and subsequently utilized as class probability. However, the problem with the response based visual explanation methods is that they can only extract attention feature maps of the attributes already present in the image. Thus, these methods are effective only in manipulations of existing attributes, such as removing beard or changing hair color. 

To address this issue, we propose a method of identifying the regions of attributes that are not present in an input image, via a novel concept of Complementary Attention FEature (CAFE). With the idea of creating a complementary attribute vector, CAFE identifies the regions to be transformed according to the input attributes even when the input image lacks the specified attributes. With CAFE, our discriminator can generate and exploit the spatial attention feature maps of all attributes. We will demonstrate CAFE’s effectiveness both in local as well as in global attributes in generating plausible and realistic images. 

Our contributions are as follows:
\begin{itemize}
    \item We present a novel approach for facial attribute editing designed to only edit the parts of a face pertinent to the target attributes based on the proposed concept of Complementary Attention FEature (CAFE).
    \item We introduce a complementary feature matching loss designed to aid in training the generator for synthesizing images with given attributes rendered accurately and in appropriate facial regions.
    \item We demonstrate effectiveness of CAFE in both local as well as global attribute transformation with both qualitative and quantitative results.
\end{itemize}

\section{Related Work}
\subsubsection{Generative Adversarial Networks.} Since Goodfellow et al.~\cite{goodfellow2014generative} proposed Generative Adversarial Network (GAN), GAN-based generative models have attracted significant attention because of their realistic output. 
However, the original formulation of GAN suffers from training instability and mode collapse. Numerous studies have been made to address these problems in various ways such as formulating alternative objective function~\cite{Arjovsky2017WassersteinG,gulrajani2017improved,Mao_2017} or developing modified architecture~\cite{karras2017progressive,karras2019style,radford2015unsupervised}. 
Several conditional methods~\cite{Isola_2017_CVPR,mirza2014conditional} have extended GAN to image-to-image translation. CycleGAN~\cite{Zhu_2017} proposed the use of cycle consistency loss to overcome the lack of paired training data. Advancing from single domain transfer methods, StarGAN~\cite{Choi_2018} can handle image translation among multiple domains. 
These developments enabled GANs to deliver some remarkable results in various tasks such as style transfer~\cite{chang2018pairedcyclegan}, super-resolution~\cite{lai2017deep,ledig2017photo}, and many other real-world applications~\cite{ak2019attribute,park2020fusion,pumarola2018ganimation}.

\subsubsection{Face Attribute Editing.} The goal of face attribute editing is to transform the input facial image according to a given set of target attributes. Several methods have been proposed with Deep Feature Interpolation (DFI) scheme~\cite{Chen_2018,chen2019semantic,upchurch2017deep}. By shifting deep features of an input image with a certain direction of target attributes, a decoder takes interpolated features and outputs an edited image. Although they produce some impressive results, limitation of these methods is that they require a pre-trained encoder such as VGG network~\cite{simonyan2014very} besides they have a weakness in multi-attribute editing. Recently, GAN based frameworks have become the dominant form of face attribute manipulation. A slew of studies for single attribute editing~\cite{kim2017unsupervised,li2016deep,liu2017unsupervised,Shen_2017,Zhang_2018_ECCV,Zhu_2017} have been conducted. However these methods cannot handle manipulation of multiple attributes with a unified model. Several efforts~\cite{lample2017fader,perarnau2016invertible} have been extended to an arbitrary attribute editing but they achieved limited quality of image. Several methods ~\cite{Choi_2018,He2017AttGANFA,zhao2018modular} have shown remarkable results in altering multiple attributes by taking the target attribute vector as an input of their generator or adopting additional network. 
STGAN~\cite{Liu_2019_CVPR} and RelGAN~\cite{wu2019relgan} further improved face editing ability by using difference between a target and a source vector to constrain in addressing selected attributes.
However these methods still suffer from change of irrelevant regions. SaGAN~\cite{Zhang_2018_ECCV} exploits spatial attention to solve the problem, but it is only effective for editing local attributes like adding mustache.
\subsubsection{Interpreting CNN.}
Several researches~\cite{fukui2019attention,selvaraju2017grad,smilkov2017smoothgrad,springenberg2014striving,zhou2016learning}  have visualized the decision making of CNN by highlighting important region. Gradient-based visual explanation methods~\cite{selvaraju2017grad,smilkov2017smoothgrad,zhou2016learning} have been widely used because they are applicable to pre-trained models. Nonetheless, these methods are inappropriate for providing spatial information to our discriminator because they require back propagation to obtain attention maps and are not trainable jointly with a discriminator. In addition to gradient-based methods, several response based methods~\cite{fukui2019attention,zhou2016learning} have been proposed for visual explanation. They obtain attention map using only response of feed forward propagation. ABN~\cite{fukui2019attention} combines visual explanation and attention mechanism by introducing an attention branch. Therefore, we adopt ABN to guide attention features in our model. However, there is a problem when applying ABN in our discriminator because it can visualize only attributes present in the input image. We combine ABN and arbitrary face attribute editing by introducing the concept of complementary attention feature to address the difficulty of localizing facial regions when the input image doesn't contain the target attribute.

\begin{figure*}[!t]
\centering \includegraphics[width=\textwidth]{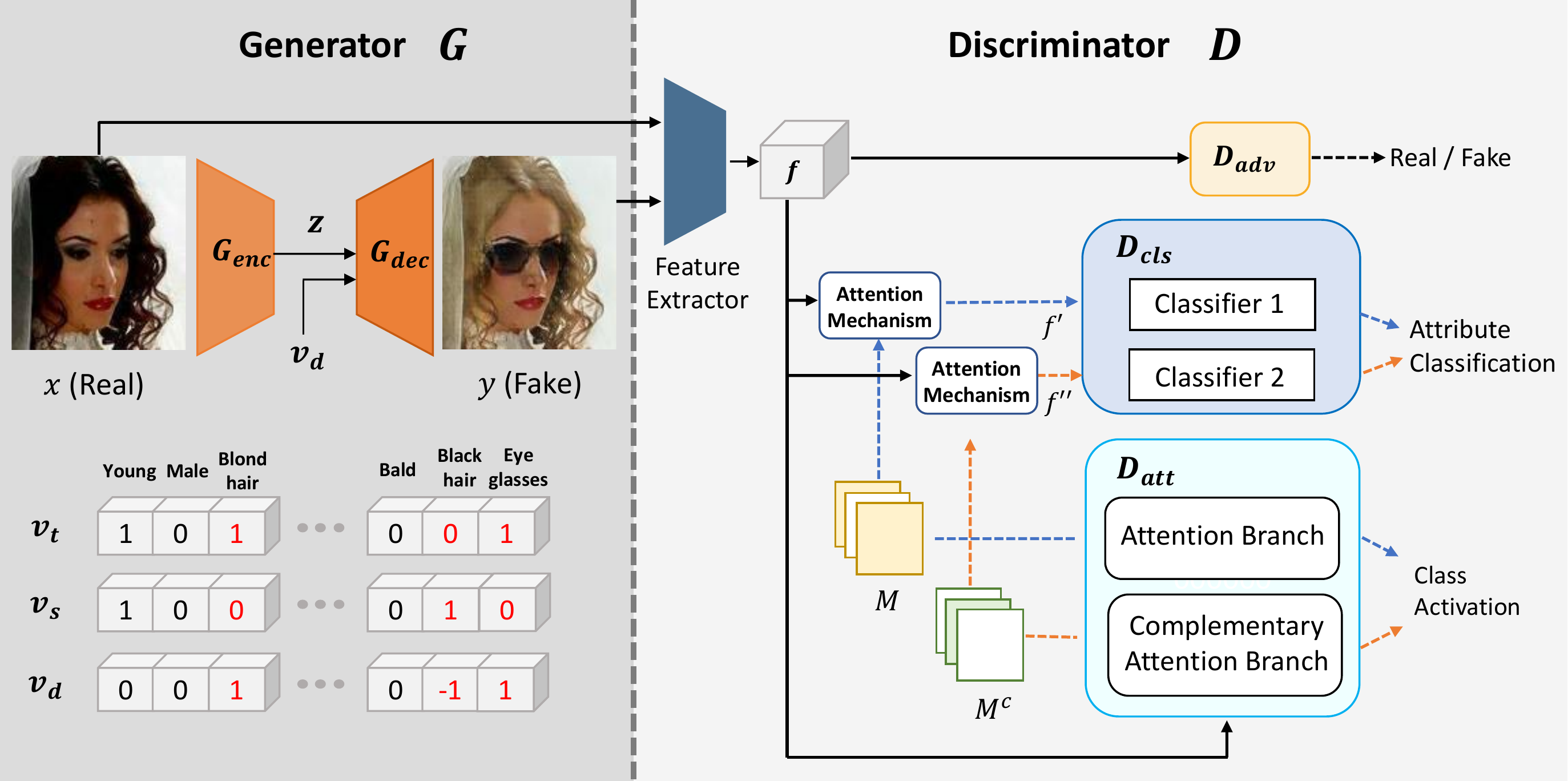}
\caption{Overview of our model. On the left side is the generator $G$ which edits source image according to given target attribute vector. $G$ consists of $G_{enc}$ and $G_{dec}$. Then the discriminator $D$ takes both source and edited image as input. $D$ consists of feature extractor, $D_{att}, D_{adv}$ and $D_{cls} $.}
\label{fig:f2}
\end{figure*}  

\section{CAFE-GAN}
This section presents our proposed CAFE-GAN, a framework to address arbitrary image editing. Fig.~\ref{fig:f2} shows an overview of our model which consists of a generator $G$ and a discriminator $D$. We first describe our discriminator that recognizes attribute-relevant regions by introducing the concept of complementary attention feature (CAFE), and then describe how the generator learns to reflect the spatial information by describing complementary feature matching. Finally, we present our training implementation in detail.

\subsection{Discriminator} 
The discriminator $D$ takes both real images and fake images modified by $G$ as input.
$D$ consists of three main parts, i.e., $D_{att}$,  $D_{adv}$, and $D_{cls}$  as illustrated in right side of Fig.~\ref{fig:f2}.
Unlike other arbitrary attribute editing methods~\cite{Choi_2018,He2017AttGANFA,Liu_2019_CVPR}, a spatial attention mechanism is applied to mid-level features $f$ in our discriminator. 
$D_{att}$ plays a major role in applying attention mechanism with a novel concept of adopting complementary feature maps. 
$D_{att}$ consists of an attention branch (AB) and a complementary attention branch (CAB) and they generate $k$ attention maps, which are the number of attributes, respectively. 
$M= \{M_1,...,M_k\}$, a collection of attention maps from AB, contains important regions of attributes that exist in an input image while $M^c=\{M^c_{1},...,M^c_{k}\}$ from CAB contains casual region of attributes that do not exist. 
These attention maps are applied to mid-level features by the attention mechanism as
\begin{align}
     f'_{i} & = f\cdot M_i, \label{eq:1}\\
    f''_{i}  &= f\cdot M^{c}_{i}, \label{eq:2} 
\end{align}
where $M_i$ and $M^{c}_{i}$ are the $i$-th attention map from AB and CAB respectively and $(\cdot)$ denotes element-wise multiplication.

\begin{figure}[!t]
\centering \includegraphics[width=\linewidth]{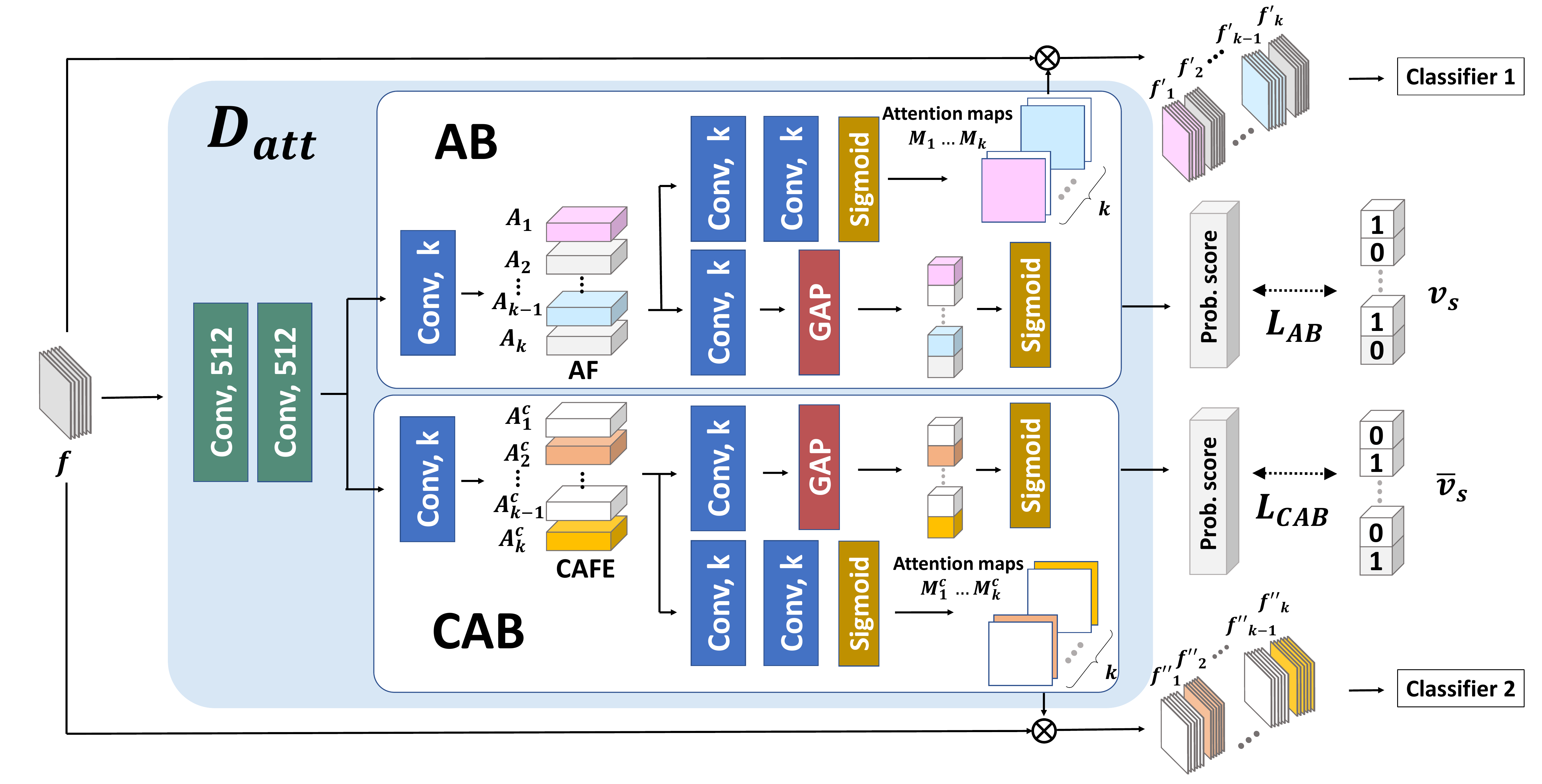}
\caption{Details on structure of attention branch (AB) and complementary attention branch (CAB) in $D_{att}$.}
\label{fig:f3}
\end{figure}

The following paragraph describes $D_{att}$ in detail as illustrated in Fig.~\ref{fig:f3}.  As explained above, we adopt Attention Branch (AB) to identify attribute-relevant regions following ABN~\cite{fukui2019attention}.
AB takes mid-level features of an input image and generate $~h~\times~w\times~k~$ attention features (AF), denoted by $A$, with a $~1~\times~1\times~k~$ convolution layer. Here, $k$ denotes the number of channels in $A$ which is the same as the number of attributes. $h$ and $w$ denote the height and width of the feature map respectively.
AB outputs $k$ attention maps $M_1,...M_k$ with $~1~\times~1\times~k~$ convolution layers and a sigmoid layer. 
It also outputs activation of each attribute class by global average pooling (GAP)~\cite{DBLP:journals/corr/LinCY13}. 
The $~h~\times~w\times~k~$ attention feature map $A$ is converted to $~1~\times~1\times~k~$ feature map by GAP to produce probability score of each class with a sigmoid layer. 
The probability score is compared to label $v_s$ with a cross-entropy loss to train $D$ to minimize classification errors when a real image (source image) is given as an input. Therefore, attention loss of AB is defined as

\begin{equation}
    \mathcal{L}_{D_{AB}} = -\mathbb{E}_{x}\sum_{i=1}^{k}\left[v_{s}^{(i)}\log{D}_{AB}^{(i)}(x)\right.+ \left.(1-v_{s}^{(i)})\log(1-D_{AB}^{(i)}(x))\right],
\end{equation}
where $x$ is real image, $v_{s}^{(i)}$ denotes the $i$-th value of source attribute vector and $D_{AB}^{(i)}(x)$ denotes the $i$-th probability score that is output of AB. 
Therefore, the values of each channel in $A$ are directly related to activation of the corresponding attribute. AB can extract $A$ which represents spatial information about attributes contained in the input image. However $A$ does not include the information about attributes that are not present in the image because $A_i$, the $i$-th channel of feature map $A$,  does not have response if $i$-th attribute is not in the input image.

This aspect does not influence the classification models like ABN because it only needs to activate a channel that corresponds to the correct attribute shown in an input image. However, for handling any arbitrary attribute, the generative model needs to be able to expect related spatial region even when an input image does not possess the attribute. Hence, there is a limit to apply existing visual explanation methods into the discriminator of attribute editing model directly. 

To address this problem, we developed the concept of complementary attention and implemented the idea in our algorithm by integrating Complementary Attention Branch (CAB).The concept of CAB is intuitive that it extracts complementary attention feature (CAFE), denoted by $A^c$, which represents causative region of attribute that are not present in image. For example, CAB detects the lower part of face about attribute $Beard$ if the beard is not in an input image. To achieve this inverse class activation, we exploit complementary attribute vector $\bar{v_{s}}$ to compare with probability score from $A^c$ and  $\bar{v_{s}}$ is formulated by
\begin{equation}
   \bar{v_{s}}=1-v_{s},
\end{equation}
hence the attention loss of CAB is formulated as 
\begin{equation}
    \mathcal{L}_{D_{CAB}} = -\mathbb{E}_{x}\sum_{i=1}^{k}\left[(1-v_{s}^{(i)})\log{D}_{CAB}^{(i)}(x)\right.\\
                        + \left.v_{s}^{(i)}\log(1-D_{CAB}^{(i)}(x))\right], 
\end{equation}
where, $D_{CAB}^{(i)}(x)$ denotes the $i$-th probability score that is output of CAB.
CAB is designed to generate a set of attention maps $M^c$ for attention mechanism from $A^c$. Therefore $A^c$ should contain spatial information to help $D_{cls}$ classify attributes. In other words, $A^c$ represents causative regions of non-existing attribute.
With AB and CAB, our model extracts attention feature map about all attributes because $A$ and $A^c$ are complementary. In other words, for any $i$-th attribute, if $A_i$ does not have response values, $A_{i}^{c}$ has them and vice versa.

Two groups of attention maps $M$ and $M^c$, outputs of AB and CAB respectively, have different activation corresponding to attributes. 
In other words, $M$ is about existing attributes of the input image while $M^c$ is about absent attributes of that. After attention mechanism, the transformed features are forwarded to two multi-attribute classifiers in $D_{cls}$ and the classifier 1 and classifer 2 classify correct label of image with $f'$ and $f''$ respectively. Each classifier outputs the probability of each attribute with cross-entropy loss. For discriminator, it learns to classify real image $x$ with two different attention mechanism, i.e.,
\begin{equation}
    \mathcal{L}_{D_{cls}} = -\mathbb{E}_{x}\sum_{n=1,2}\sum_{i=1}^{k}\left[v_{s}^{(i)}\log{D}_{cls_{n}}^{(i)}(x)\right.
                        + \left.(1-v_{s}^{(i)})\log(1-D_{cls_{n}}^{(i)}(x))\right],
\end{equation}
where $D_{cls_1}$ and $D_{cls_2}$ stand for two classifiers using collections of attention maps $M=\{M_1,...,M_k\}$ and $M^c=\{M_{1}^{c},...,M_{k}^{c}\}$, respectively. Therefore, the reason CAFE can represent the spatial information of non-existent attributes is that CAB has to generate the attention maps that can help to improve performance of the classifiers while reacting to non-existent attributes by GAP.

In $D$, there is another branch $D_{adv}$ distinguishes real image $x$ and fake image $y$ in order to guarantee visually realistic output with adversarial learning. In particular, we employ adversarial loss in WGAN-GP~\cite{gulrajani2017improved}, hence the adversarial loss of $D$ is given as 
\begin{equation} 
    \mathcal{L}_{D_{adv}}=\mathbb{E}_{x}(D_{adv}(x))-\mathbb{E}_y(D_{adv}(y))
                         -\lambda_{gp}\mathbb{E}_{\hat{x}}\left[(\lVert{\nabla_{\hat{x}}D_{adv}(\hat{x})}\rVert_2-1)^2\right],
\end{equation}
where $\hat{x}$ is weighted sum of real and fake sample with randomly selected weight $\alpha\in[0,1]$.

\begin{figure}[!t]
\centering \includegraphics[width=\linewidth]{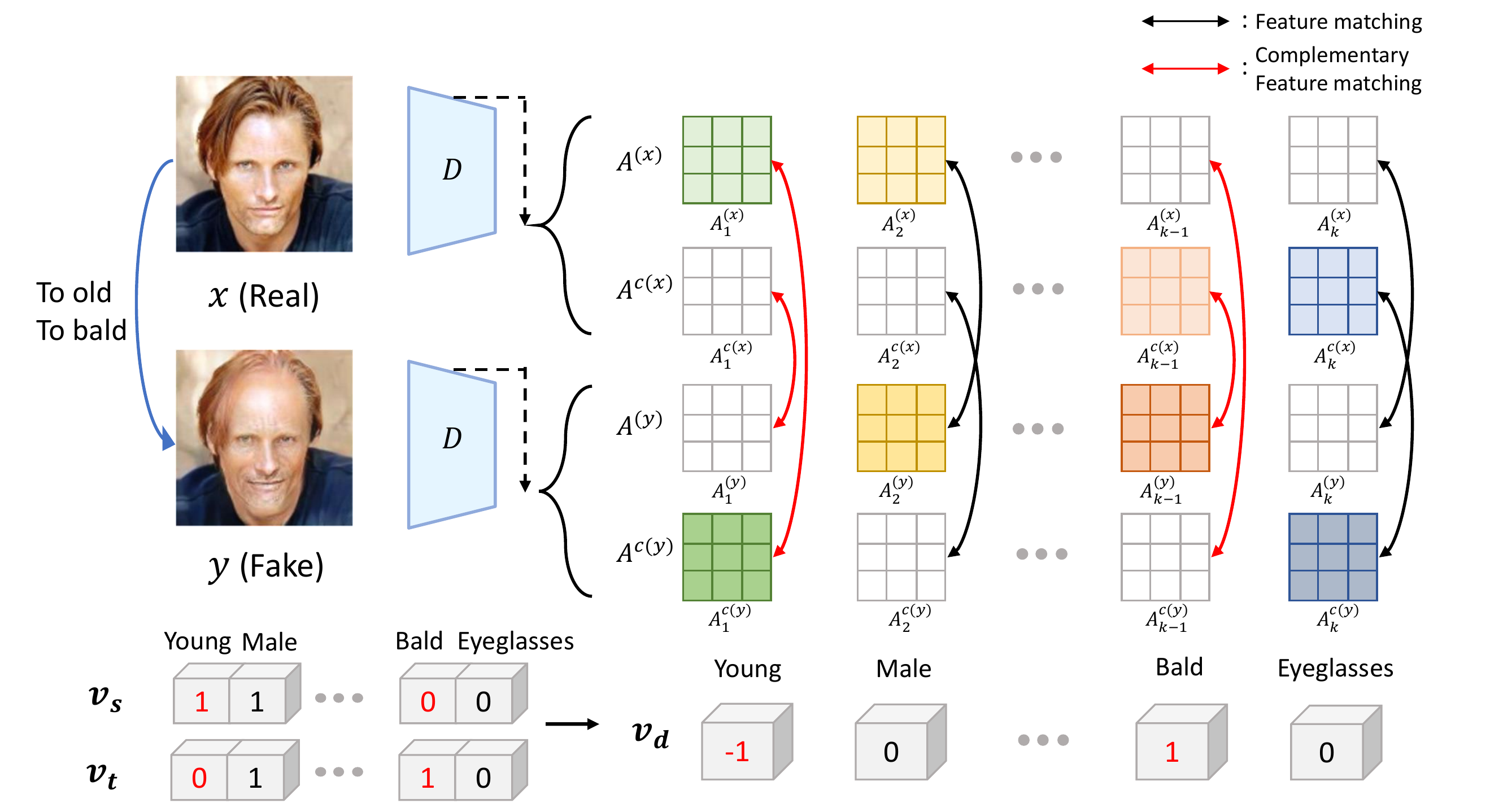}
\caption{Details on the proposed complementary feature matching with an example. $G$ learns to match AF and CAFE of edited image corresponding to attribute to be changed, i.e., \textit{To old} and \textit{To bald}, to CAFE and AF of source image, respectively. On the contrary, G learns to match AF and CAFE of edited image to AF and CAFE of source image for the attribute not to be changed. }
\label{fig:f4}
\end{figure}  

\subsection{Generator}
The generator $G$ takes both the source image $x$ and the target attribution label $v_t$ as input, and then conducts a transformation of $x$ to $y$, denoted by $y=G(x,v_t)$. The goal of $G$ is to generate an image $y$ with attributes according to $v_t$ while maintaining the identity of $x$. $G$ consists of two components: an encoder $G_{enc}$ and a decoder $G_{dec}$. From the given source image $x$, $G_{enc}$ encodes the image into a latent representation $z$. Then, the decoder generates a fake image $y$ with the latent feature $z$ and the target attribute vector $v_t$. Following ~\cite{Liu_2019_CVPR}, we compute a difference attribute vector $v_d$ between the source and the target attribute vectors and use it as an input to our decoder. 
\begin{equation}
    v_{d}=v_{t}-v_{s},
\end{equation} 
Thus, the process can be expressed as
\begin{align}
    z&=G_{enc}(x),\\
    y&=G_{dec}(z, v_{d}).
\end{align} 
In addition, we adopt the skip connection methodology used in STGAN~\cite{Liu_2019_CVPR} between $G_{enc}$ and $G_{dec}$ to minimize loss of fine-scale information due to down sampling. After that, $D$ takes both the source image $x$ and the edited image $y$ as input. $G$ aims to generate an image that can be classified by $D_{cls_1}$ and $D_{cls_2}$ as target attribute, hence the classification loss of $G$ is defined as 
\begin{equation}
    \mathcal{L}_{G_{cls}} = -\mathbb{E}_{y}\sum_{n=1,2}\sum_{i=1}^{k}\left[v_{t}^{(i)}\log{D}_{cls_{n}}^{(i)}(y)\right.
                        + \left.(1-v_{t}^{(i)})\log(1-D_{cls_{n}}^{(i)}(y))\right],
\label{eq:eq7}
\end{equation}
where $v_{t}^{(i)}$ denotes the $i$-th value of target attribute vector.

Although $D_{att}$ in our discriminator can obtain the spatial information about all attributes, it is necessary to ensure that $G$ has ability to change the relevant regions of given target attributes.
Attention feature maps of the source image and the edited image should be different on those corresponding to attributes that are changed while the rest of attention feature maps should be the same. 
Therefore we propose a novel complementary matching method as illustrated in Fig.~\ref{fig:f4}. 
For the attributes that are not to be changed, the attention feature maps of edited image should be same with that of source image. 
In other words, $G$ learns to match  AF of edited image to AF of source image when the attributes remain the same (black arrows in Fig.~\ref{fig:f4}), and $G$ also learns from matching CAFE of the source image. 
When the given target attributes are different from the source image, $G$ learns to match  AF of edited image to CAFE of source image (red arrows in Fig.~\ref{fig:f4}), same with CAFE of edited image to AF of source image.
Let $\{A^{(x)},A^{c(x)}\}$ and $\{A^{(y)},A^{c(y)}\}$ denote set of AF and CAFE from two different samples, real image x and fake image y, respectively.
Complementary matching is conducted for changed attributes and thus the complementary matching loss is defined as    
\begin{align}
\begin{split}
    \mathcal{L}_{CM}=\mathbb{E}_{(x,y)}\sum_{i=1}^{k}\frac{1}{N_i}[
    \lVert{A_{i}^{(x)}-P_{i}}\rVert_{1} + \lVert{A_{i}^{c(x)}-Q_{i}}\rVert_{1}],\\
    \text{where} \ \{P_{i},Q_{i}\}= \begin{cases}\{A_{i}^{(y)},A_{i}^{c(y)}\}&\text{if}\ |v_{d}^{(i)}|=0,\\
                                                    \{A_{i}^{c(y)},A_{i}^{(y)}\}&\text{if}\ |v_{d}^{(i)}|=1,
    \end{cases}
    \end{split}
\end{align}
where $k$ is the number of attributes and $N(i)$ denotes the number of elements in feature map. $A_{i}^{(x)/(y)}$ and $A_{i}^{c(x)/(y)}$ denote $i$-th channel of $A^{(x)/(y)}$ and $A^{c(x)/(y)}$ respectively and  $v_{d}^{(i)}$ denotes $i$-th value of difference attribute vector $v_d$.  

For adversarial training of GAN, we also adopt the adversarial loss to generator used in  WGAN-GP~\cite{gulrajani2017improved}, i.e.,  
\begin{equation}
    \mathcal{L}_{G_{adv}}=\mathbb{E}_{x,v_{d}}[D_{adv}(G(x,v_{d}))], 
\end{equation}

Although the generator can edit face attribute with $\mathcal{L}_{{G}_{cls}}$ and generate realistic image with $\mathcal{L}_{{G}_{adv}}$, it should preserve identity of image. Therefore, $G$ should reconstruct the source image when difference attribute vector is zero. We adopt the pixel-level reconstruction loss, i.e., 

\begin{equation} \label{eq:eq14}
    \mathcal{L}_{rec}=\mathbb{E}_{x}[\lVert{x-G(x,\mathbf{0})}\rVert_1],
\end{equation}
where we use $\ell_1$ loss for sharpness of reconstructed image and $\mathbf{0}$ denotes zero vector.
\subsection{Model Objective}

Finally, the full objective to train discriminator $D$ is formulated as
\begin{equation} \label{eq:eq15}
        \mathcal{L}_{D}=  -\mathcal{L}_{D_{adv}}+\lambda_{att}\mathcal{L}_{D_{att}}
        +\lambda_{D_{cls}}\mathcal{L}_{D_{cls}},
\end{equation}
and that for the generator $G$ is formulated as 
\begin{equation} \label{eq:eq16}
    \mathcal{L}_{G} =  \mathcal{L}_{G_{adv}}+\lambda_{CM}\mathcal{L}_{CM}+\lambda_{G_{cls}}\mathcal{L}_{G_{cls}}+\lambda_{rec}\mathcal{L}_{rec},
\end{equation}
where $\lambda_{att}, \lambda_{D_{cls}}, \lambda_{CM}, \lambda_{G_{cls}}$ and $\lambda_{rec}$ are hyper-parameters which control the relative importance of the terms.

\section{Experiments}
In this section, we first explain our experimental setup and then present qualitative and quantitative comparisons of our model with the state-of-the-art models. Finally, we demonstrate effectiveness of CAFE with visualization results and ablation study. The experiments not included in this paper can be found in supplementary material.

\subsection{Experimental Setup}
We use CelebFaces Attributes (CelebA) dataset~\cite{liu2015deep} which consists of 202,599 facial images of celebrities. Each image is annotated with 40 binary attribute labels and cropped to $178\times218$. We crop each image to $170\times170$ and resize to $128\times128$. For comparison, we choose the same attributes used in the state-of-the-art models~\cite{Choi_2018,He2017AttGANFA,Liu_2019_CVPR}.   
In our experiment, coefficients of the objective functions in Eq.~(\ref{eq:eq15})  and (\ref{eq:eq16}) are set to $\lambda_{att}=\lambda_{D_{cls}}=\lambda_{CM}=1,\lambda_{G_{cls}}=10,$ and $\lambda_{rec}=100$. We adopt ADAM~\cite{kingma2014adam} solver with $\beta_1=0.5$ and $\beta_2=0.999$, and the learning rate is set initially to 0.0002 and set to decay to 0.0001 after 100 epochs.

\begin{figure}[!t]
\centering \includegraphics[width=\linewidth]{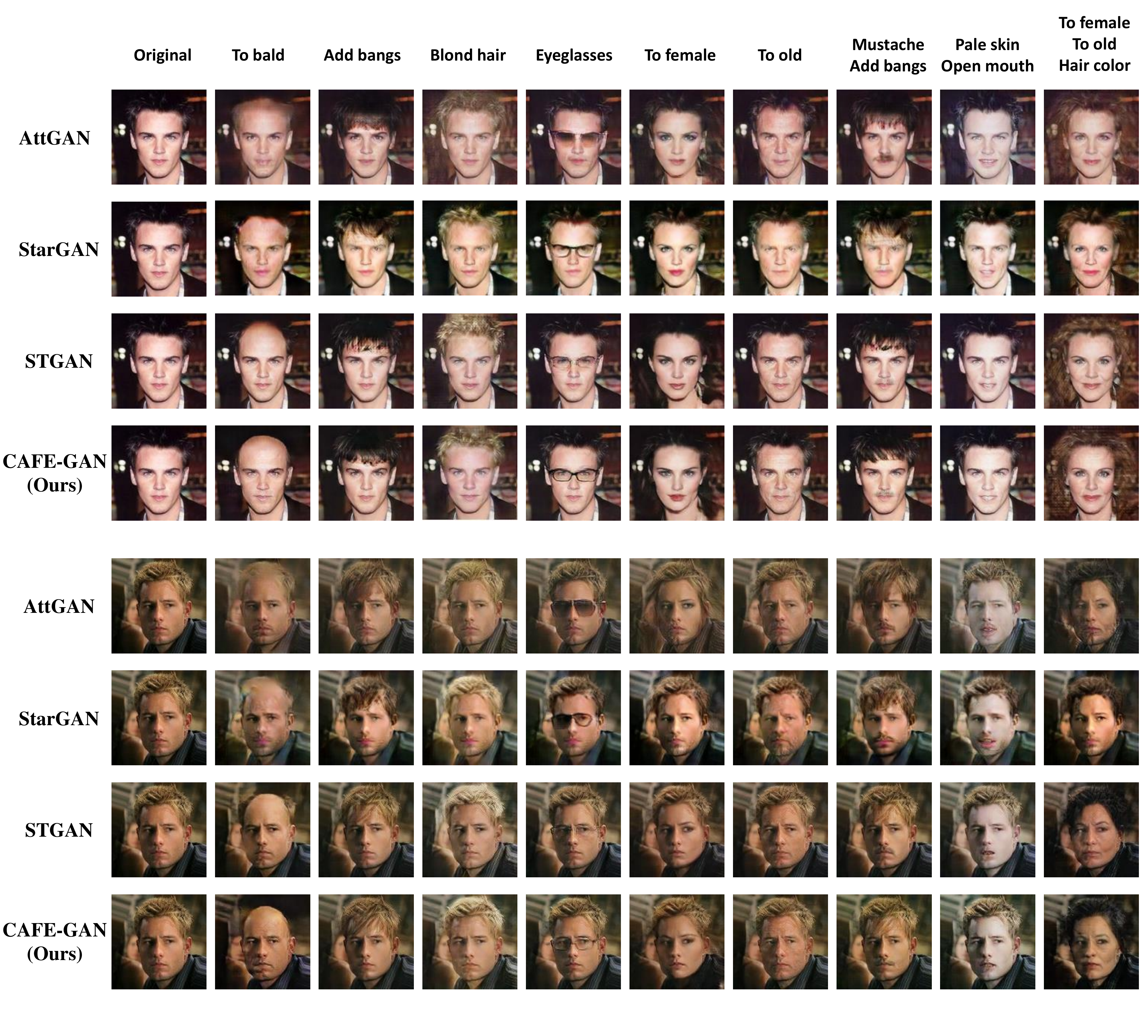}
\caption{Qualitative comparison with arbitrary facial attribute editing models. The row from top to down are result of AttGAN~\cite{He2017AttGANFA}, StarGAN~\cite{Choi_2018}, STGAN~\cite{Liu_2019_CVPR} and our model. \textbf{Please zoom in to see more details.}}
\label{fig:f5}
\end{figure}  

\subsection{Qualitative Result}
    
First, we conduct qualitative analysis by comparing our approach with three state-of-the-art methods in arbitrary face editing, i.e., AttGAN~\cite{He2017AttGANFA}, StarGAN~\cite{Choi_2018} and STGAN~\cite{Liu_2019_CVPR}. The results are shown in Fig.~\ref{fig:f5}. 
Each column represents different attribute manipulation and each row represents qualitative results from the methods compared. The source image is placed in the leftmost of each row and we analyze results about single attribute as well as multiple attributes. First, AttGAN~\cite{He2017AttGANFA} and StarGAN~\cite{Choi_2018} perform reasonably on attributes such as \textit{Add bangs}. However they tend to edit irrelevant regions for attributes such as \textit{Blond hair} or \textit{Pale skin} and they also result in blurry images for attributes such as \textit{To bald}. STGAN~\cite{Liu_2019_CVPR} improves manipulating ability by modifying structure of the generator, but this model also presents unnatural images for some attributes like \textit{To Bald} and \textit{Add Bangs}. In addition, unwanted changes are inevitable for some attributes such as \textit{Blond hair}. As shown in Fig.~\ref{fig:f5}, our model can successfully convert local attributes like \textit{Eyeglasses} as well as global attributes like \textit{To female} and \textit{To old}. Last three columns represent results of multi-attribute editing. \textit{Hair color} in the last column denotes $\textit{Black}~\leftrightarrow~\textit{Brown hair}$.  It can be seen that our model delivers more natural and well-edited images than the other approaches.

\begin{figure}[!t]
\centering \includegraphics[width=\linewidth]{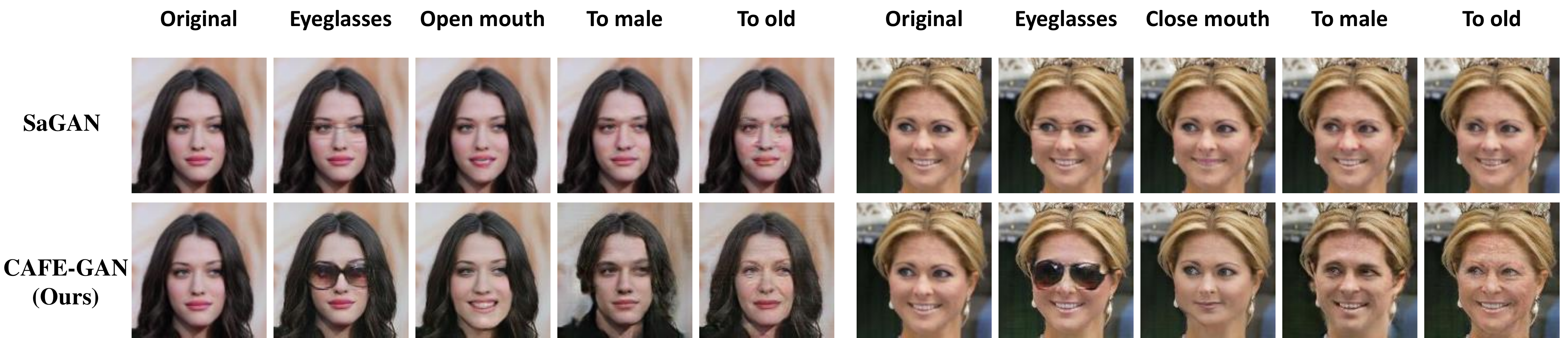}
\caption{Qualitative comparison with SaGAN~\cite{Zhang_2018_ECCV} which adopt spatial attention in the generator. \textbf{Please zoom in to see more details.}}
\label{fig:f6}
\end{figure}  

We also compare our model with SaGAN~\cite{Zhang_2018_ECCV} which adopt spatial attention concept. As shown in Fig.~\ref{fig:f6}, SaGAN could not edit well in global attributes like \textit{To male} and \textit{To old}. Even when the target attribute is on a localized region like \textit{Eyeglasses}, it performed poorly. However, our model calculates the spatial information in feature-level, not in pixel-level. As a result, our model shows well-edited and natural results in both local and global attributes. In addition, note that SaGAN is a single-attribute editing model, wherein it requires one model to be trained per attribute. 

\begin{table}[t] \centering
\caption{Comparisons on the attribute classification accuracy. Numbers indicate the classification accuracy on each attribute.}
\label{tb1:accuracy}
\begin{tabular}{p{1.8cm}p{1.0cm}p{1.0cm}p{1.2cm}p{1.0cm}p{1.1cm}p{1.0cm}p{0.9cm}p{1.3cm}p{1.0cm}}
\toprule
            & Bald    & Bangs     & Blond h.  & Musta. & Gender & Pale s. & Aged & Open m. & Avg.\\
\midrule
AttGAN~\cite{He2017AttGANFA}  &23.67 &91.08 &41.51 &21.78 &82.85 &86.28 &65.69 &96.91 &63.72 \\
STGAN~\cite{Liu_2019_CVPR}   &59.76 &95.48 &79.98  &\textbf{42.10} &92.70 &97.81 &85.86 &\textbf{98.65} &81.54 \\     
CAFE-GAN    &\textbf{79.03} &\textbf{98.59} &\textbf{88.14} &40.13 &\textbf{95.22} &\textbf{98.20} &\textbf{88.61} &97.15 &\textbf{85.64} \\

\bottomrule

\end{tabular}
\end{table}

\subsection{Quantitative Result}\label{Sec:sec4.3}
We present quantitative results by comparison with two of the three models compared earlier~\cite{He2017AttGANFA,Liu_2019_CVPR}.
In the absence of the ground-truth, the success of arbitrary attribute editing can be determined by a well trained multi-label classifier, i.e., well edited image would be classified to target domain by a well-trained attribute classifier.
For fair comparison, we adopted the classifier that was used to evaluate attribute generation score for STGAN~\cite{Liu_2019_CVPR} and AttGAN~\cite{He2017AttGANFA}.To evaluate quantitative result, we use official code which contains network architecture and weights of parameter in well-trained model. We exclude StarGAN~\cite{Choi_2018} in this section because the official code of StarGAN provides only few attributes.
There are 2,000 edited images evaluated for each attribute and their source images come from the test set in CelebA dataset.
We measure the classification score of each attribute of edited images and they are listed in Table~\ref{tb1:accuracy}. In Table~\ref{tb1:accuracy}, \enquote{Blond h.} and \enquote{Open m.} denote \textit{Blond hair} and \textit{Open mouth} respectively.
While our model shows competitive scores on the average for many attributes, it also delivered overwhelming results compared to the other models for specific attributes such as \textit{Bald}. For attributes such as \textit{Mustache} and \textit{Open mouth}, STGAN results in slightly better performance over our model.   

\subsection{Analysis of CAFE}
This section presents analysis of the proposed method. First we show the result of visualization of our attention features and then we conduct ablation study to demonstrate effectiveness of CAFE.

\subsubsection{Visualization of CAFE.}
\begin{figure}[t]
\centering \includegraphics[width=\linewidth]{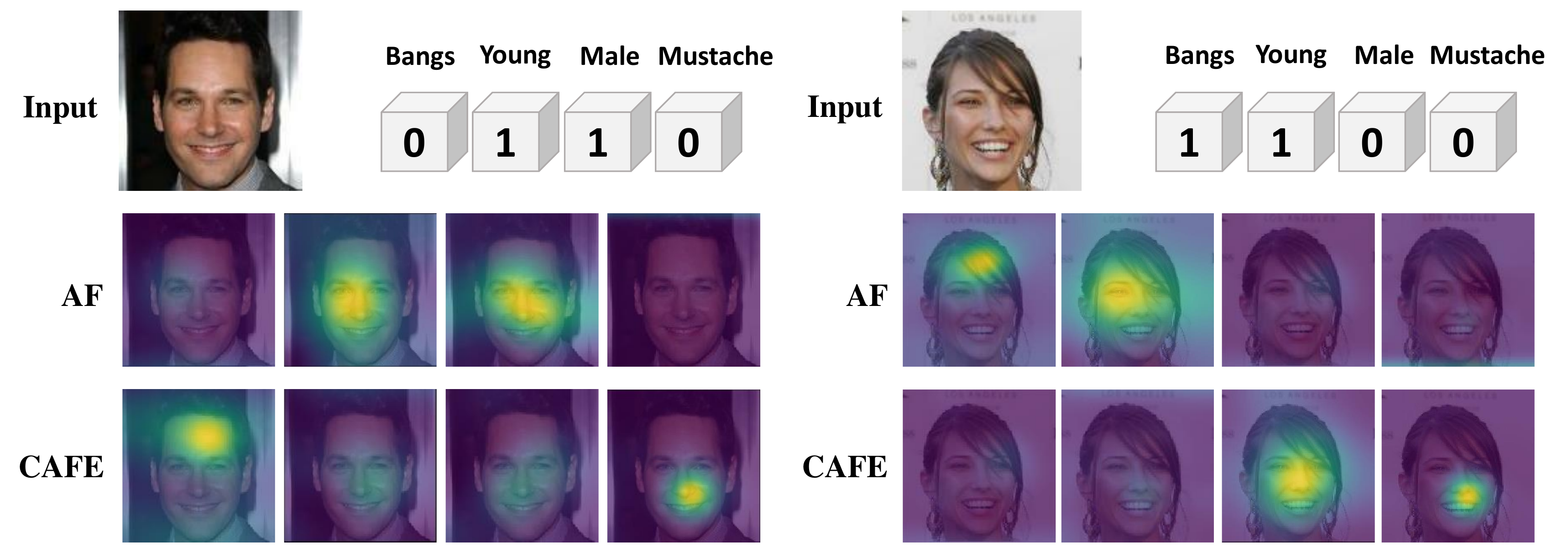}
\caption{Visualization results of the attention features (AF) and the complementary attention features (CAFE).}
\label{fig:f7}
\end{figure}

Fig.~\ref{fig:f7} shows visualization results of AF and CAFE to examine whether they activate related region correctly. Because the man in left in Fig.~\ref{fig:f7} has no bangs, AF rarely activates but CAFE activates and highlights the related region correctly. The result on the right shows AF correctly activating the region relevant to bangs while CAFE doesn't. Since CAFE only finds complement features absent in the specified attributes, not activating the region specified in the attribute is the correct response. The remainder of the figure demonstrates that CAFE lights up the regions complementary to the given attributes accurately. For global attributes like \textit{Young} and \textit{Male}, both AF and CAFE respond correctly. Although AF and CAFE does not detect attribute-relevant regions at pixel level since they are considered at feature-level, they highlight the corresponding regions accurately per given attribute. As such, our model performs better on editing both global and local attributes compared with other methods.

\subsubsection{Ablation Study.}
\begin{figure}[!t]
\centering \includegraphics[width=\linewidth]{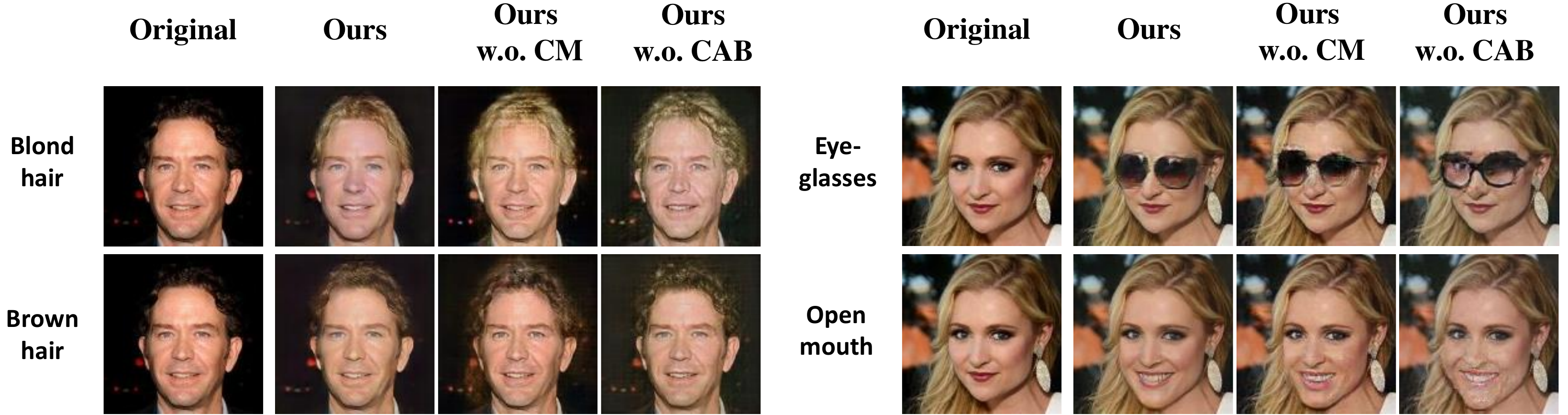}
\caption{Qualitative results of CAFE-GAN variants.}
\label{fig:f8}
\end{figure}

To demonstrate effectiveness of the proposed method, we evaluate performance of our model by excluding key components one by one.
We compare our model with two different versions, i.e., (i) CAFE-GAN w.o. CM : excluding complementary matching loss ($\mathcal{L}_{CM}$) in training process and (ii) CAFE-GAN w.o. CAB : removing complementary attention branch (CAB) in our discriminator.
As shown in Fig.~\ref{fig:f8}, the generator has a difficulty to determine where to change without complementary matching loss though the discriminator can extract AF and CAFE. Some results from the model without CM show unwanted and over-emphasized changes. Excluding CAB leads to artifacts and unwanted changes in generated images because the discriminator has limited spatial information about the attributes that are contained in the input image
We also measure classification accuracy of each attribute by the pre-trained classifier which was used in Section~\ref{Sec:sec4.3} and the results are listed in Table~\ref{tb2:abl_accuracy}. 
In the absence of CM or CAB, the classification accuracy decreases in all attributes except \textit{Open mouth}, and the model without CAB shows the lowest accuracy.

\section{Conclusion}
We introduced the CAFE-GAN based on the complementary attention features and attention mechanism for facial attribute editing.  The CAFE controls facial editing to occur only on parts of the facial image pertinent to specified target attributes by exploiting the discriminator's ability to locate spatial regions germane to specified attributes. Performance of CAFE-GAN was compared with the state-of-the-art methods via qualitative and quantitative study.  The proposed approach in most of the target attributes demonstrated improved performances over the state-of-the-art methods, and in some attributes achieved significantly enhanced results.

\begin{table}[t] \centering
\caption{Attribute classification accuracy of ablation study.}
\label{tb2:abl_accuracy}
\begin{tabular}{p{1.8cm}p{1.0cm}p{1.0cm}p{1.2cm}p{1.0cm}p{1.1cm}p{1.0cm}p{0.9cm}p{1.3cm}p{1.0cm}}
\toprule
            & Bald    & Bangs     & Blond h.  & Musta. & Gender & Pale s. & Aged & Open m. & Avg.\\
\midrule
Ours  &\textbf{79.03} &\textbf{98.59} &\textbf{88.14} &\textbf{40.13} &\textbf{95.22} &\textbf{98.20} &\textbf{88.61} &97.15 &\textbf{85.64} \\
w.o. CM   &61.68 &97.46 & 87.01  &39.78 &85.93 &92.38 &86.39 &\textbf{97.56}&81.02  \\     
w.o. CAB   &32.43 &91.45 &70.41 &36.13 &81.93 &92.22 &65.95 & 94.72 & 70.66  \\

\bottomrule

\end{tabular}
\end{table}

\noindent\textbf{Acknowledgment} Authors (Jeong-gi Kwak and Hanseok Ko) of Korea University are supported by a National Research Foundation (NRF) grant funded by the MSIP of Korea (number 2019R1A2C2009480). David Han’s contribution is supported by the US Army Research Laboratory.


\clearpage
%
%

\bibliographystyle{splncs04}
\bibliography{egbib}
\end{document}


\pagestyle{headings}
\mainmatter
\def\ECCVSubNumber{2202}  

\title{CAFE-GAN: Arbitrary Face Attribute Editing\\ with Complementary Attention Feature\\ -Supplementary material-} 

\author{Jeong-gi Kwak\inst{1} \and
David K. Han\inst{2} \and
Hanseok Ko\inst{1}}
\authorrunning{J. Kwak et al.}
\titlerunning{CAFE-GAN}
%
\institute{Korea University, Seoul, Korea \and
Army Research Laboratory, Adelphi, MD, USA}

\maketitle

\setcounter{section}{5}
\section{Appendix}

In this section, we supplement additional quantitative and qualitative results that are not reported in the main paper. 
To demonstrate effectiveness of our methods in terms of visual quality and preventing undesired change, we first report two quantitative results i.e., FID score (subsection~\ref{sec:sec6.1}) and user study (subsection~\ref{sec:sec6.2}) with state-of-the-art methods~\cite{Choi_2018,He2017AttGANFA,Liu_2019_CVPR} 
In addition, we present additional comparison results with~\cite{Choi_2018,He2017AttGANFA,Liu_2019_CVPR,wu2019relgan} in subsection~\ref{sec:sec6.3} and then show extra results of CAFE-GAN in subsection~\ref{sec:sec6.4}.

\subsection{FID Score}\label{sec:sec6.1}
We first report FID (Fréchet Inception Distance) scores~\cite{heusel2017gans} which is commonly used as metrics of GAN models. 
We use the following methods and conditions to measure FID scores.
First, we split the celebA test set in two subsets (TestA and TestB). TestA is used as source image of editing task and TestB is used to measure FID score. Therefore, we compare two sets of images (Test A$’$(set of fake images resulted from TestA) vs.~TestB(real)). 
By doing this, we try to avoid the effect of similar input and output when measuring FID scores and the results are listed in Table.~\ref{sptb1:fid_score}. Our model outperforms the other methods in all tasked attributes except for a few cases. 
Note that we set every image to undergo some changes (e.g., bang $\rightarrow$ no bang, no bang $\rightarrow$ bang if \textit{Bangs} is given), so that an evaluation by FID score becomes meaningful. 

\begin{table}[b] \centering
\caption{FID (Fréchet Inception Distance) scores. Lower is better. }
\label{sptb1:fid_score}
\begin{tabular}{p{2.0cm}p{1.0cm}p{1.0cm}p{1.2cm}p{1.0cm}p{1.0cm}p{1.0cm}p{1.0cm}p{1.0cm}p{1.0cm}}
\toprule
            & Bald    & Bangs     & Blond h.  & Eyegl. & Gender & Musta.  & Aged & Mouth & Avg.\\
\midrule
AttGAN  &18.51 &14.38 &18.11 &34.89 &22.35 &9.52 &13.02 &9.11 &17.49 \\
StarGAN &25.15 &17.62 &25.16 &26.71 &20.59 &16.85 &18.83 &13.67 &20.57  \\     
STGAN   &14.97 &8.03 &15.28 &12.83 &11.01 &\textbf{4.99} &7.03 & 6.33&10.06  \\
CAFE-GAN  &\textbf{13.73} &\textbf{7.85} &\textbf{14.00} &\textbf{8.05} &\textbf{8.10} &5.47 &\textbf{5.59} & \textbf{4.84}&\textbf{8.44}  \\
Real img. & & & & & & & & &3.03  \\

\bottomrule

\end{tabular}
\end{table}

\subsection{User Study}\label{sec:sec6.2}

Since there is no ground truth as to where to change or preserve in attribute editing, the criteria for this can vary from person to person. We thus report results of a user study with a survey platform. The number of attributes used in the evaluation is six. Each user evaluated 10 images per attribute, thus a total of 60 images were evaluated by each user, and they were directed to choose the best edited image with quality considering the specified attributes. The results are listed in Table~\ref{sptb2:user_study}. Our model achieves higher score that other models in five attributes and with a large gap in specific attributes such as \textit{Bald}. Our model get lower score in \textit{Bangs}. 

\begin{table}[t] \centering
\caption{Comparisons on the user preference. Numbers indicate
the percentage of preference on each attribute. }
\label{sptb2:user_study}
\begin{tabular}{p{2.0cm}p{1.2cm}p{1.2cm}p{1.4cm}p{1.2cm}p{1.2cm}p{1.9cm}p{1.0cm}}
\toprule
            & Bald    & Bangs     & Blond h.  & Gender   & Aged & Mouth & Avg.\\
\midrule
AttGAN  &1.34 &\textbf{29.34} &13.42 &13.2 &14.37 &15.29 &14.49 \\
StarGAN &2.39 &21.73 &11.19 &5.26 &13.64 &4.64 &9.81   \\     
STGAN   &21.42 &22.56 &24.32 &35.21 &31.56 &12.45 &24.59  \\
CAFE-GAN  &\textbf{74.85} &26.37 &\textbf{51.07} &\textbf{46.33} &\textbf{40.43} &\textbf{67.62} &\textbf{51.11}  \\

\bottomrule

\end{tabular}
\end{table}

\subsection{Additional Comparison Results}\label{sec:sec6.3}
\begin{figure}
\centering \includegraphics[width=\linewidth]{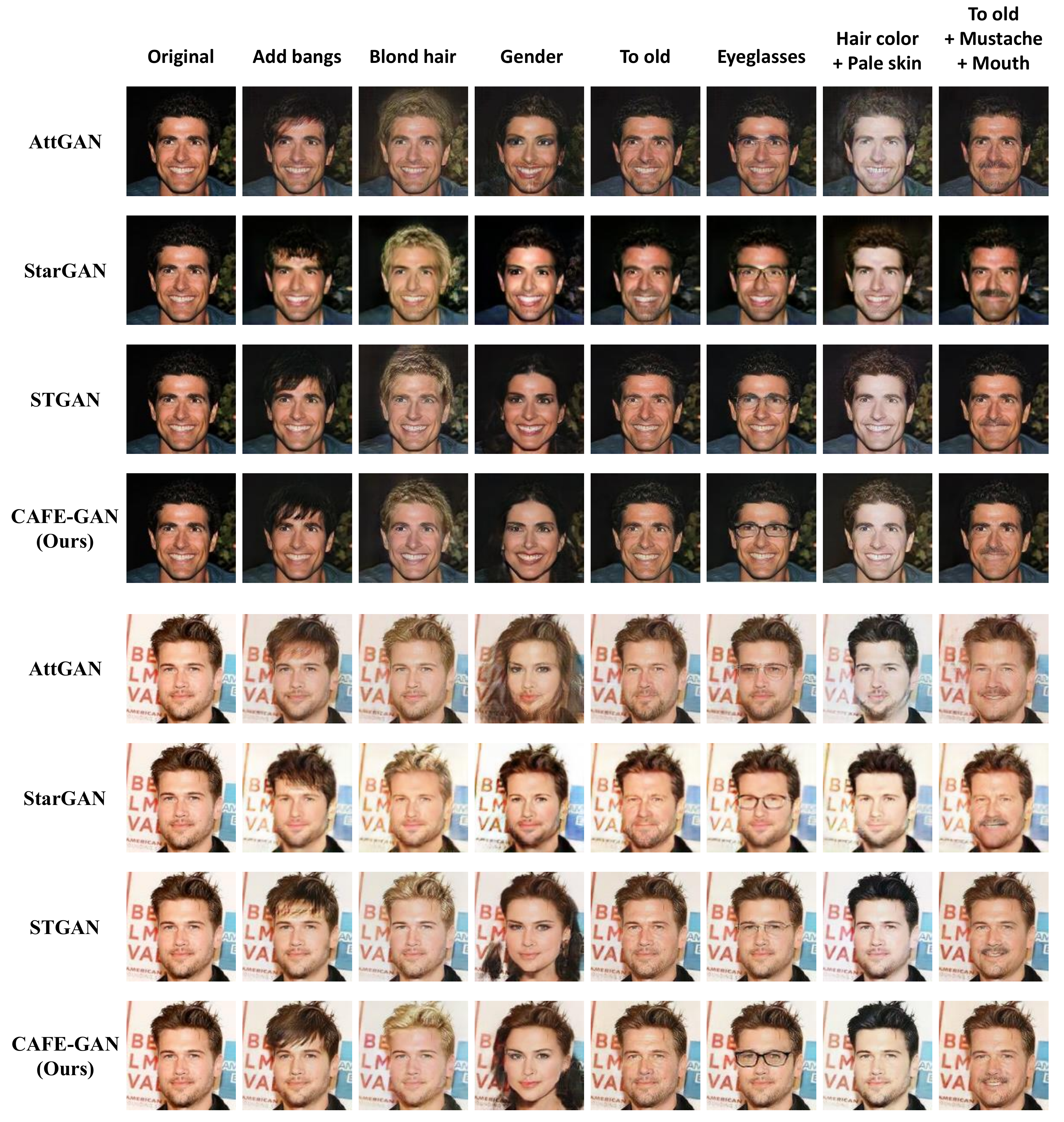}
\caption{Additional results of qualitative comparison on CelebA ($128\times128$) images.}
\label{fig:sf1}
\end{figure}

\begin{figure}[!t]
\centering \includegraphics[width=\linewidth]{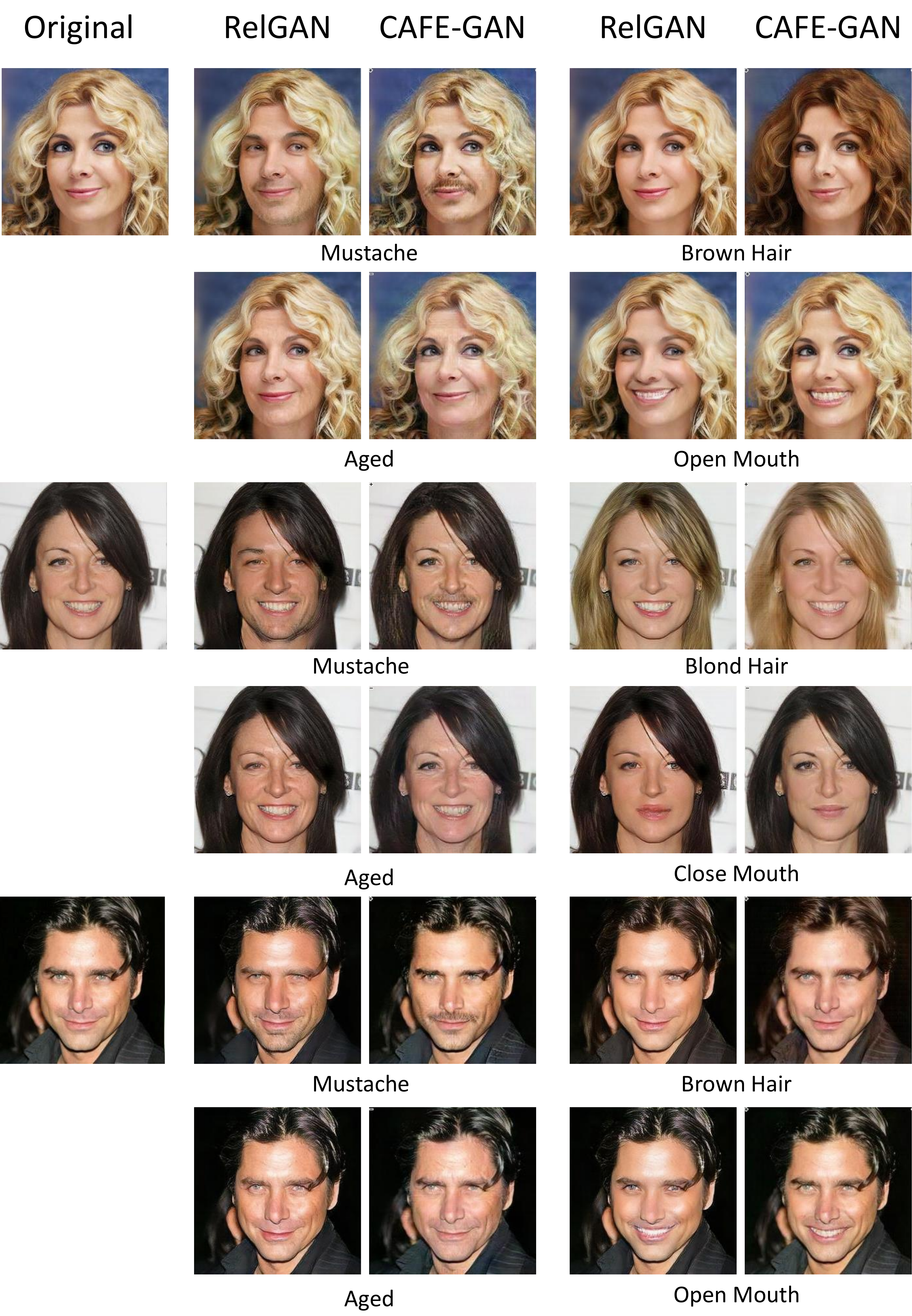}
\caption{Qualitative comparison with RelGAN on CelebA-HQ ($256\times256$) images.}
\label{fig:sf2}
\end{figure}

 We first present additional comparison of qualitative results with same models used in main paper, i.e., AttGAN~\cite{He2017AttGANFA}, StarGAN~\cite{Choi_2018}, and STGAN~\cite{Liu_2019_CVPR} on CelebA~\cite{liu2015deep} images ($128\times128$) in Fig.~\ref{fig:sf1}. we also present comparison results on CelebA-HQ dataset~\cite{karras2017progressive} which contains 30,000 images in CelebA with high quality. We use $256\times256$ images for comparison with RelGAN~\cite{wu2019relgan} which use the difference between a target and a source vector (relative attributes) to constrain in addressing selected attributes. The results of RelGAN have some unintended changes in edited images. For example, when \textit{Mustache} is given as an attribute to be inserted, RelGAN edits female look to like male as shown in Fig.~\ref{fig:sf2}, but CAFE-GAN edits only the regions related to mustache. Furthermore, CAFE-GAN expresses fine details better when \textit{Open} or \textit{Close Mouth} is given. 


\subsection{Extra Results of CAFE-GAN}\label{sec:sec6.4}

\begin{figure}[!t]
\centering \includegraphics[width=\linewidth]{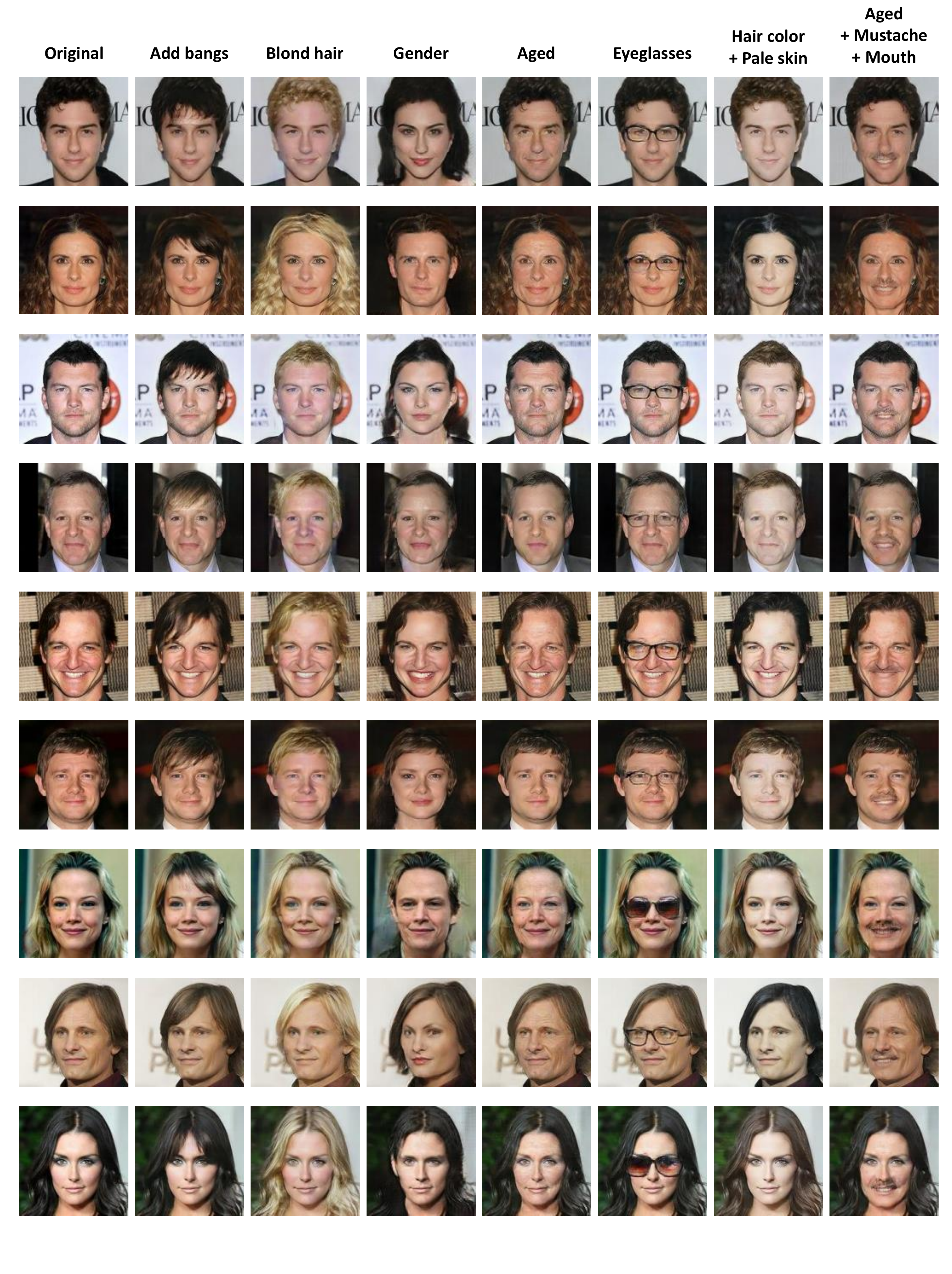}
\caption{Extra qualitative results of our model. }
\label{fig:sf3}
\end{figure}

Finally, we present extra qualitative results of CAFE-GAN in Fig.~\ref{fig:sf3}. Through the results, we demonstrate a superior visual quality and editing performance of our model.

\clearpage

\bibliographystyle{splncs04}
\bibliography{egbib}